\documentclass[conference]{IEEEtran}
\IEEEoverridecommandlockouts
\usepackage{cite}
\usepackage{amsmath,amssymb,amsfonts}
\usepackage{algorithmic}
\usepackage{graphicx}
\usepackage{textcomp}
\usepackage{xcolor}
\usepackage{booktabs}

\def\BibTeX{{\rm B\kern-.05em{\sc i\kern-.025em b}\kern-.08em
    T\kern-.1667em\lower.7ex\hbox{E}\kern-.125emX}}
\begin{document}

\title{Leveraging Previous Facial Action Units Knowledge for Emotion Recognition on Faces}

\author{\IEEEauthorblockN{Pietro B. S. Masur}
\IEEEauthorblockA{\textit{Voxar Labs - Centro de Informática} \\
\textit{Universidade Federal de Pernambuco}\\
Recife, Brazil \\
pbsm@cin.ufpe.br}
\and
\IEEEauthorblockN{Willams Costa}
\IEEEauthorblockA{\textit{Voxar Labs - Centro de Informática} \\
\textit{Universidade Federal de Pernambuco}\\
Recife, Brazil \\
wlc2@cin.ufpe.br}
\and
\IEEEauthorblockN{Lucas S. Figueredo}
\IEEEauthorblockA{\textit{Unidade Acadêmica de Belo Jardim} \\
\textit{Universidade Federal Rural de Pernambuco}\\
Belo Jardim, Brazil \\
lsf@cin.ufpe.br}
\and
\IEEEauthorblockN{Veronica Teichrieb}
\IEEEauthorblockA{\textit{Voxar Labs - Centro de Informática} \\
\textit{Universidade Federal de Pernambuco}\\
Recife, Brazil \\
vt@cin.ufpe.br }

}

\IEEEoverridecommandlockouts
\IEEEpubid{\makebox[\columnwidth]{979-8-3503-4807-1/23/\$31.00~\copyright2023 IEEE\hfill} \hspace{\columnsep}\makebox[\columnwidth]{ }}

\maketitle

\IEEEpubidadjcol
\begin{abstract}
   People naturally understand emotions, thus permitting a machine to do the same could open new paths for human-computer interaction. Facial expressions can be very useful for emotion recognition techniques, as these are the biggest transmitters of non-verbal cues capable of being correlated with emotions. Several techniques are based on Convolutional Neural Networks (CNNs) to extract information in a machine learning process. However, simple CNNs are not always sufficient to locate points of interest on the face that can be correlated with emotions. In this work, we intend to expand the capacity of emotion recognition techniques by proposing the usage of Facial Action Units (AUs) recognition techniques to recognize emotions. This recognition will be based on the Facial Action Coding System (FACS) and computed by a machine learning system. In particular, our method expands over EmotiRAM, an approach for multi-cue emotion recognition, in which we improve over their facial encoding module.
\end{abstract}


\section{Introduction}
\label{sec:intro}

Emotions are an inherent aspect of a human being's life. They are an indicator of one's inner state, which is constantly communicated through verbal and nonverbal cues in one's communication. The ability to recognize what is being expressed by an individual is vital for communication to happen. In particular, reading someone's emotional state is essential to correspond with them in the correct tone. For example, when talking to someone sad, people act differently than when talking to someone happy.

Being able to classify the emotions of a user during the interaction with an intelligent system or environment leads to generating useful data for decision-making. Social networks and streaming platforms currently ask the user for their opinion regarding content to generate better recommendations in the future \cite{fadhelaljunid2017survey}. However, asking for the input of a user may not always be the best option; users may reply only if they dislike too much or like too much. For some applications, such as, for example, education, safety, healthcare, and entertainment, having insights into the user's perceived emotion are not only helpful but also enabling \cite{vinola15}. Many components go into building a robust solution for emotion recognition; for instance, the metrics and outputs from such a model must be consistent with psychological principles. 

When it comes to performing recognition of emotions on faces, another fundamental of psychology can be applied: the Facial action encoding system (FACS) \cite{facs}. FACS is a system capable of translating someone's non-verbal cues into an emotion. It was coded based on empirical observation of subjects' facial and corporal displays while feeling different emotions.

\begin{figure}[!h]
	\centering {\includegraphics[scale=0.53]{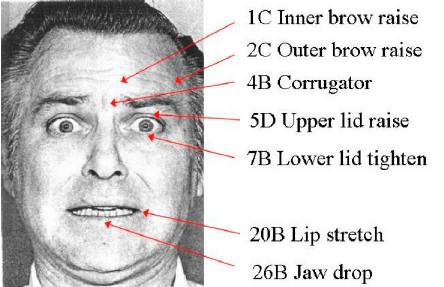}}
	\hfill
	\caption{The FACS encoding of fear, from  \cite{littlewort}.}
	\label{fear}
\end{figure}

FACS encodes specifically the behaviors behind the activation of AUs; those associated with facial muscles are referred to as facial action units (FAUs). When someone feels an emotion, his body and face respond to that feeling in the form of action units \cite{facialexpr} \cite{unobservablefacial}. The response intensity varies according to each subject's physiology, but the correlation between emotions and action unit activation is universal. Besides, since this activation is a physiological response, it cannot be posed consistently. Fig. \ref{fear} displays some of the AUs activated while the subject feels fear.

Therefore, given the correlation of FAUs and emotions, in this work, we propose to evaluate if deep learning models trained to recognize FAUs could provide a more efficient face encoding than those deployed currently in emotion recognition techniques. We evaluate our work in the well-known CAER-S dataset \cite{caers} and compare it with a state-of-the-art approach for emotion recognition called EmotiRAM \cite{emotiram}, specifically against its face encoding stream, which we will refer to as EmotiRAM-f. We also propose experiments to discuss if such models could lead to better explainability of predictions by highlighting which facial cues were used for the prediction.



\section{Related Works}

\textit{\textbf{Emotion recognition.}} Over the years, techniques for emotion recognition have been proposed based on using more than one nonverbal cue to identify emotion; mainly context representation, which contains descriptions of the context that the person is placed into. The \textit{Context-aware emotion 
 recognition networks} (CAER-Net) \cite{caers} is one of the first works to explore this possibility while also publishing the CAER and CAER-S datasets and proposing a public benchmark. The CAER-S dataset was generated by extracting videos and frames from 79 television shows. The main performer would always be somewhat aligned with the camera, and the frame would have contextual information. For CAER-Net, the authors propose two main encoding streams, namely the face encoding stream and the context encoding stream, in which information from each cue would flow into and have features extracted. Although this approach led to significant results, there were still possibilities for improvement regarding the approach for cue extraction.

 Later, in work named \textit{Emotion recognition on adaptive multi-cues} (EmotiRAM) \cite{emotiram}, the authors propose to extend this pipeline by improving the extraction of representations from context and by proposing an independent stream to process body language. Each cue is processed individually and their representations are fused at the end of the inference. Although proposing a significant improvement, which also led to a significant accuracy, both EmotiRAM and CAER-Net lack a face encoding stream that is capable of extracting deep representations linked to emotion since these methods only search for generic cues in the facial regions.

 \textit{\textbf{Facial action units activation.}} A possible way to extend these works is related to how FAUs are extracted by the current state-of-the-art of this field. Sanchez \textit{et al.} \cite{sanchez_et_al} proposes in their work a supervised learning approach for learning AU heatmaps. Their proposed methodology is to estimate the facial landmarks and generate Gaussian maps around the points where AUs are known to cause changes. The inspiration for the approach is related to joint activations of AUs. Thus, it is possible to model AUs more realistically by considering them all when performing regression. The authors employ an Hourglass network \cite{hourglass_network} as a backbone of their model, which achieved significant results.

However, relying on pre-defined rules for modeling AU co-occurrence leads to limited generalization. Each face contains a unique setting; taking this into consideration provides a fuller approach to this problem. To contour that, Fan \textit{et al.} proposes a novel GNN-based technique \cite{pytorch-fau}. Their framework is such that latent relationships of AUs are automatically learned via establishing semantic correspondences between feature maps. This is achieved by employing a combination of heatmap regression and Semantic Correspondence Convolution (SCC). SCC modules are based on the work by Wang \textit{et al.} \cite{wang2019dynamic}, which focuses on dynamic graph convolutions in geometry modeling. Intuitively, feature channels that are simultaneously activated are considered to have a latent co-relation; this corresponds to a co-occurrence pattern of AU intensities.

The method basic framework is built by adding several de-convolutional layers on a ResNet \cite{resnet_article}, which generates the AUs feature maps. Feature maps are then input into SCC modules 3 times. The network outputs one heatmap per AU. 

Finally, considering each facial topology when building AU graphs is essential for building a robust GNN-based technique. However, the number of edges between the AUs may also be a very important factor in the representation ability of this graph. Luo \textit{et al.} \cite{Luo_2022} propose that a single edge between AUs in a graph representation is insufficient for dealing with their complex relations. To address this problem, they propose a strategy that, for each AU pair, encodes a pair (two edges) of multi-dimensional features.

To achieve their objective and deal with the problems they've encountered with previous approaches to model AUs relationships as graphs, the authors utilize three resources:
\begin{enumerate}
    \item Modeling a full face representation into an explicitly AUs relationship describing graph through the Attention Node Feature Learning (ANFL) module
    \item Modeling the relationship between AUs pair on a multi-dimensional edge feature graph with Multi-edge Feature Learning (MEFL) module
    \item Considering  the uniqueness of each facial display by utilizing full face representation as input to the two modules listed above
\end{enumerate}

The ANFL module is composed of two blocks: 
\begin{enumerate}
    \item AU-specific Feature Generator (AFG): composed of Fully Connected (FC) and global average polling layers which jointly act as a feature extractor 
    \item Facial Graph Generator (FGG): computes  node similarity via KNN and GCN. This block is only used as a reinforcement to the AFG block; thus it is not used on training 
\end{enumerate}

MEFL module also has two blocks: 
\begin{enumerate}
    \item Facial Display-specific AU Representation Modeling (FAM) Receives full face representation and facial features from the AFG FC layer and locates activation cues of each AU 
    \item AU Relationship Modeling (ARM)
\end{enumerate}
 
 The ARM outputs are utilized to extract features from the located cues which relate to both AUs activations.

\section{Method}

Our proposal with this investigation is to improve the accuracy of EmotiRAM-f \cite{emotiram} through the knowledge of FAUs. To achieve this, we propose using models trained for this task as a baseline and retraining this model on the CAER-S dataset to perform emotion recognition tasks. 

Ekman and Friesen \cite{facs} disclose how emotions can be recognized based on the AUs activation, which is the basis for this evaluation. Furthermore, the emotion displayed by the AUs are non-posed; thus, AUs consists a better source for emotion inference than naive facial observation. Leveraging this observation in constructing this approach, we hypothesize that encoding action units consist of a better source of information than naive facial encoding techniques. This is the postulate we depart from to develop our EmotiRAM-FAU technique. 



Therefore, given an image \(I\), we aim to classify this image into a perceived emotion \(y\) from a list of possible emotions \(C \in \mathbb{R}^{7}\). For this, we extend the pipeline proposed by Fan \textit{et al.} \cite{pytorch-fau} by adding bottlenecks to manage the flow of representations and the generation of the classification. Following the original implementation pipeline, we use a ResNet \cite{resnet_article} as a backbone and add de-convolutional layers to upsample the filters and generate the heatmaps. After this, we add a bottleneck sequence using Fully Connected (FC) layers to transcribe this internal representation into a classification \(C\) for the given image. The heatmap generated by the pre-trained model has a shape of \(10\times24\times24\). This is a fixed shape that is restrained by the output of Fan \textit{et al.} architecture. The shape is flattened  to a resulting \(5760\times1\) shape and passed through the following arbitrated FC layers. We reduce the shape of this internal representation using a base-2 approach with 5 FC layers as following: \(5760\)-dim; \(2048\)-dim; \(1024\)-dim; \(512\)-dim; \(256\)-dim and finally \(7\)-dim.




\begin{figure*}
	\centering {\includegraphics[width=0.8\linewidth]{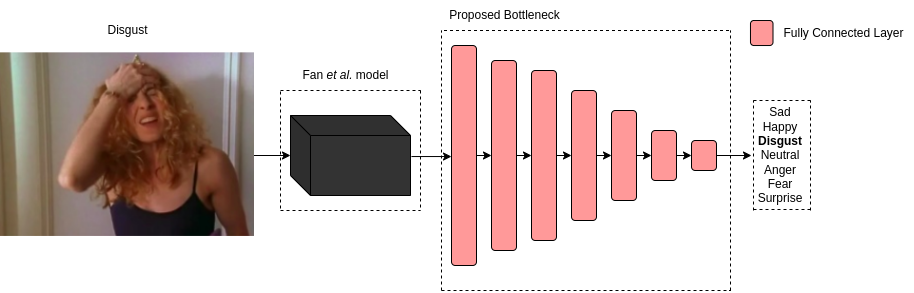}}
	\hfill
	\caption{Graphical representation of our architecture. We treat Fan \textit{et al.} model as a black box and link it with a bottleneck, represented in red. Then the bottleneck outputs the value corresponding to the emotion it has predicted.}
	\label{fig:emotiram-fau-arch}
\end{figure*}


\section{Experiments}

\textit{\textbf{Datasets.}} For this experimentation, we use a dataset for the task of FAU and another dataset for the emotion recognition task. For FAU we employ DISFA \cite{disfa}, a dataset built from the collection of videos from twenty-seven young adults. The participants were recorded while videos aiming to induce spontaneous expressions were shown to them. The recordings were made with two cameras, one on the right and the other one on the left side. The data collected was labeled by FACS experts, frame by frame, to indicate the presence (or absence) and intensity of FAUs.  DISFA contains labels for 12 AUs: 5 on the upper face and 7 on the lower face. In joint with BP4D\footnote{This is another dataset that could fit well in our experimentation; however, we could not gain access to it. Therefore, we use models pre-trained on them.} \cite{bp4d}, DISFA is a default benchmark for AUs recognition techniques.

The CAER benchmark \cite{caers} comprises videos collected from 79 TV shows, totaling 20,484 clips which were manually annotated with six different emotion labels. CAER-S is a static version of the CAER benchmark. On this version, there is a single frame from the original CAER benchmark video which is annotated in the same way as the videoclip.

To perform our experiments, we have split CAER-S into three partitions: train, validation, and test. We used pre-trained models for the task of facial action unit activation as a baseline in which we implemented our approach. Our approach consists of inserting a bottleneck of FC layers into the pre-trained model and retraining them to recognize emotions. The bottleneck outputs represent the detected emotions. 

\textit{\textbf{Environment and implementation.}} We reimplemented EmotiRAM-f as discussed on Costa \textit{et al.} \cite{emotiram} and trained on CAER-S using PyTorch 1.11.0. For the experiments related to FAUs, we reuse the source code published by Luo \textit{et al.} \cite{Luo_2022} \footnote{https://github.com/EvelynFan/FAU}, utilizing PyTorch 1.4.0, as stated in their public repository. We perform our experiments on a desktop computer running Ubuntu 20.04 with an RTX 2080 Ti GPU with 24 GB of RAM, and a 4-core processor with 8 threads.

\subsection{Experimental Settings}

We have experimented with three different approaches based on ResNet-50 for FAU recognition. First, the work by Fan \textit{et al.}  \cite{pytorch-fau}, in which we used a pre-trained model in the BP4D dataset, and also the work by Luo \textit{et al.}  \cite{Luo_2022} with a pre-trained model in the BP4D dataset and another pre-trained model in the DISFA dataset. Luo \textit{et al.} model will be further referred to as ME-graph.

\begin{enumerate}
    \item Fan \textit{et al.} \cite{pytorch-fau} model pre-trained on BP4D
    \item Luo \textit{et al.} \cite{Luo_2022} model pre-trained on DISFA
    \item Luo \textit{et al.} \cite{Luo_2022} model pre-trained on BP4D
\end{enumerate}

Each of these models is separately inserted on EmotiRAM's architecture in joint with a bottleneck of FC layers, it functions as a replacement for the face-encoding module. The fine-tuned models are then evaluated in the CAER-S dataset for emotion recognition.  Luo \textit{et al.} technique output is a vector of probabilities for each AU on the dataset. Thus, in our experiments with it, only one FC layer could be utilized.

\subsection{Experiments on Explainability}

Given that we use pre-trained models for the task of AU recognition, we propose an experiment to assess the correlation between the internal representation of AUs and those fine-tuned for emotions, despite no mechanism that controls which representations to keep
being utilized. The pre-trained models we use give output information about AUs. In particular, Luo \textit{et al.} model outputs a vector of probabilities of the activations of each AU. For this experiment, we only utilize ME-graph models pre-trained on DISFA since it is the only AU dataset we have access to.

\subsubsection{AU experiments on DISFA}
To evaluate how much the fine-tuned model still relies on the AUs original representation, we perform experiments on the DISFA dataset. For this end, a publicly available implementation of Luo \textit{et al.} code was
utilized. First, we evaluate metrics for the original pre-trained ME-graph model. Then, we evaluate our fine-tuned model, truncating its bottleneck to fit with the network architecture required by Luo \textit{et al.} code.

\subsubsection{AU experiments on CAER-S}
In this experiment, we extract EmotiRam-FAU's output before and after the bottleneck thus retrieving two outputs: recognized AUs and recognized emotions. We work with data we have access to for building an explainable program based on EmotiRam-FAU. DISFA contains labels for a limited range of AUs; on Luo \textit{et al.} work, the training process is limited to 8 AUs. Matching the AUs covered by ME-Graph with the emotions provided by the CAER-S dataset, we observe that the only full match between activated AUs and corresponding emotion is Happiness. We label CAER-S happiness data as being (AU6+AU12) and evaluate the correctness of the AU output made by the FAU encoding module after our fine-tuning process. We also perform the same experiments with the original ME-graph pre-trained model and use it as a baseline. In the next section, we will display our achieved results and perform some qualitative evaluations over selected images of the CAER-S dataset.

\section{Results and discussion}

Here we detail the results achieved on CAER-S in terms of accuracy.  Table \ref{table:plain_CAER-S_results} shows the results we acquired with each technique.

\begin{table}[h!]
\caption{Results achieved on the CAER-S test partition, in terms of accuracy.}
\centering
\begin{tabular}{cccc}
    \toprule
           & CAER-S \\ \midrule
    EmotiRAM-f &  70\% \\
    EmotiRAM (face+context) & 87\%\\
    EmotiRAM (face+context+body) & 89\%   \\
    EmotiRAM-FAU [BP4D] &  \textbf{77\%}  \\ 
    EmotiRAM-FAU-ME-graph [BP4D] &  62\%    \\
    EmotiRAM-FAU-ME-graph [DISFA] &  66\%    \\ \bottomrule
\end{tabular}
\label{table:plain_CAER-S_results}
\end{table}

As  shown, EmotiRAM-FAU improved by 7 percent over Emoti-
RAM’s face encoding module (EmotiRAM-f) from Costa textit{et al.} \cite{emotiram} in terms of accuracy. This is a significant result and points out that models pre-trained on AUs can provide a better encoding than naive techniques. 

The results achieved with Luo \textit{et al.} technique were not as good as the others. We believe this occurred because their network outputs are shaped as AUs probabilities, while Fan \textit{et al.} works with a heatmap richer in features as their output. Thus, when coupled with a bottleneck, Luo \textit{et al.} output features have little margin to find new vector spaces representing emotions. From now on, we will refer to EmotiRAM-FAU with Luo \textit{et al.} technique as EmotiRAM-FAU-ME-Graph.

\subsection{Results on AUS}
To evaluate how our model behaved on the AUs recognition task, we performed a test on the DISFA dataset. As previously explained could not work with the BP4D dataset due to our lack of access to the original database. 

\subsubsection{Results on DISFA}
Table \ref{tab:disfa_results} display the results achieved by EmotiRAM-FAU-ME-Graph and compares them with the original Luo \textit{et al.} model (ME-Graph). 

\begin{table*}[t]
\footnotesize
\caption{Results achieved on the CAER-S test partition, in terms of accuracy and F1 score. AUs 6 and 12 are responsible for the happiness facial expression, wich will be evaluated on the next section.}

\centering
\begin{tabular}{ccccccc}
    \toprule
           & F1 Avg  & F1 AU6 & F1 AU12 & Acc Avg & Acc AU6 & Acc AU12 \\ \midrule
    EmotiRAM-FAU-ME-Graph &  18\% & 1\% & 56\% & 62\% & 91\% & 87\% \\
    ME-Graph              & 58\% & 64\% & 82\% & 94\% & 94\% & 94\% \\ \bottomrule
\end{tabular}
\label{tab:disfa_results}
\end{table*}

As we can see, the F1 score and accuracy metrics have greatly decayed compared to the original approach. There are some reasons we believe may have caused this. As previously observed, CAER-S is an on-the-wild dataset ensembled mainly from TV shows, thus, it is fair to expect some samples to contain posed emotions. Furthermore, since the annotation of CAER-S was not made by FACS experts, we believe that they are far from ideal in terms of correspondence between the real displayed emotion (based on the FACS system) and the labeled emotion. It is also reasonable to suppose that, if our hypothesis is true, there is no reliable way to determine whether CAER-S data provides a balanced distribution of FAUs activations examples. Thus, it is natural to suppose those factors lead our model to suffer some degree of forgetfulness at its original task. We believe that having more reliable training data could greatly benefit our model approach. 

\subsubsection{Qualitative Results on CAER-S}

\begin{figure}[!h]
	\centering {\includegraphics[width=0.65\linewidth]{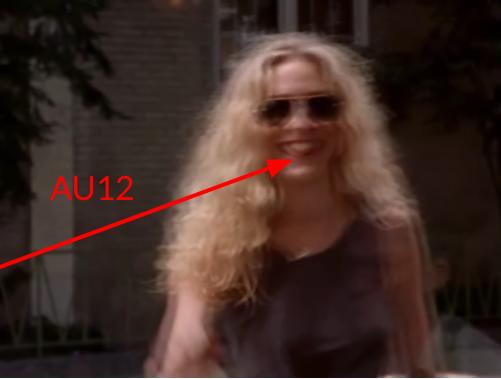}}
	\hfill
	\caption{A happiness-labeled image from the CAER-S dataset, which our system correctly labeled. Notice that AU12 is the main responsible for the happiness on this photo, and the AU6 region (eyes), in which our system has low performance, is covered by sunglasses.}
	\label{fig:happiness}
\end{figure}

\begin{figure}[!h]
	\centering {\includegraphics[width=0.65\linewidth]{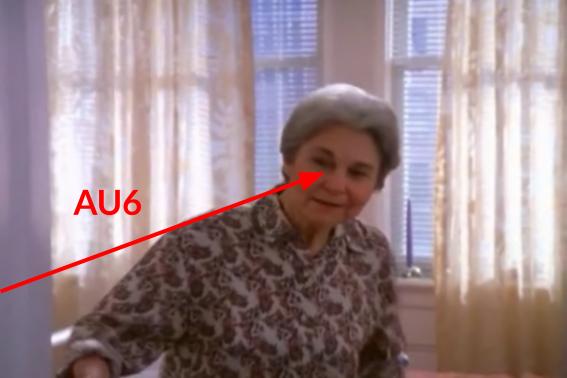}}
	\hfill
	\caption{A happiness-labeled image from the CAER-S dataset, which our system mislabeled. Notice that in this sample, the main sign of happiness is given by AU6, the same which our system practically lost the ability to recognize.}
	\label{fig:happiness_wrong}
\end{figure}

We have utilized data labeled as happiness on the CAER-S dataset and assumed that their corresponding activated AUs were AU6 and AU12. We then made a script to perform an accuracy test based on the emotion and AUs outputs produced by EmotiRAM-FAU. Table \ref{tab:CAER-S-AU} shows our achieved results. Fig \ref{fig:happiness} shows a piece of data from the CAER-s dataset, which was correctly labeled for both AUs and emotions by our approach. On the other hand, Fig. \ref{fig:happiness_wrong} displays an image our model failed to recognize as having the correct AUs.

\begin{table}[t]
\caption{For the AU6 and AU12 metrics, we consider the model is right if it can recognize the activations of both AUs, the other metric considers recognizing at least one AU as a correct answer.}
\centering
\begin{tabular}{cccc}
    \toprule
           & AU6 and AU12 & AU6 or AU12 \\ \midrule
    ME-Graph [DISFA] & 52\% & 98\% \\
    EmotiRAM-FAU-ME-Graph [DISFA] &   10\% & 92\%\\
    ME-Graph [BP4D] & 49\% & 81\% \\ \bottomrule
\end{tabular}
\label{tab:CAER-S-AU}
\end{table}

Notice that EmotiRAM-FAU-ME-Graph results at recognizing both AUs were near zero. This comes from the previously discussed forgetfulness suffered by the model. In particular, AU6 was greatly affected by this forgetfulness, as shown in table \ref{tab:disfa_results}. This is the cause of such a low score in recognizing both AUs.

Unfortunately, due to time and accessible data limitations, it was impossible to conduct more extensive and quantitative research on the effectiveness of our system. In addition, the lack of annotated data and emotions compatible with the AUs learned by the ME-Graph technique was a significant drawback to our research.

\section{Conclusion}

\subsection{Considerations}
In this work, we have investigated the use of AUs to recognize emotions in ML models. We have shown how models pre-trained to recognize AUs fine-tuned for emotion recognition tasks provide a better basis for emotion recognition than training architectures from scratch.

Our approach is explicitly aimed at emotion recognition. However, we have also investigated how much the final model intermediary representations are still faithful to the original AUs representation. Our results showed that, when fine-tuned for emotions, the model still retains some ability to recognize AUs. This is an important step towards explicable systems for emotion recognition. 

{\small
\bibliographystyle{plain}
\bibliography{egbib}
}

\end{document}